\documentclass[11pt,a4paper]{article}
\usepackage[hyperref]{acl2020}
\usepackage{times}
\usepackage{latexsym}

\usepackage{microtype}

\usepackage{graphicx}
\usepackage{amsfonts}
\usepackage{amsmath}
\usepackage{amssymb}
\usepackage{pifont}
\usepackage{nicefrac}
\usepackage{multirow}
\usepackage{subfigure}
\usepackage[font=small]{caption}
\usepackage{soul}
\usepackage{xcolor}
\usepackage{booktabs}
\usepackage{algorithm}
\usepackage{algorithmic}
\usepackage{comment}

\usepackage{tablefootnote}

\definecolor{bblue}{HTML}{4F81BD}
\definecolor{rred}{HTML}{c4260b}
\definecolor{ggreen}{HTML}{098c1f}
\definecolor{ppurple}{HTML}{9F4C7C}
\definecolor{oorange}{HTML}{F79646}

\usepackage{enumitem}
\setenumerate[1]{itemsep=0.30pt,partopsep=0pt,parsep=\parskip,topsep=5pt}
\setitemize[1]{itemsep=0.30pt,partopsep=00pt,parsep=\parskip,topsep=5pt}
\setdescription{itemsep=0.30pt,partopsep=0pt,parsep=\parskip,topsep=5pt}

\usepackage{todonotes}

\aclfinalcopy % Uncomment this line for the final submission
\newcommand\lzdrop{\textsc{${L}_0$Drop}}

\title{On Sparsifying Encoder Outputs in Sequence-to-Sequence Models}

\author{Biao Zhang$^1$ \quad Ivan Titov$^{1,2}$ \quad Rico Sennrich$^{3,1}$ \bigskip\\
  $^1$School of Informatics, University of Edinburgh \\
  $^2$ILLC, University of Amsterdam \\
  $^3$Department of Computational Linguistics, University of Zurich \\
  \texttt{B.Zhang@ed.ac.uk, ititov@inf.ed.ac.uk, sennrich@cl.uzh.ch}
  }

\date{}

\begin{document}
\maketitle
\begin{abstract}

Sequence-to-sequence models usually transfer \textit{all} encoder outputs to the decoder for generation. 
In this work, by contrast, we hypothesize that these encoder outputs can be compressed to shorten the sequence delivered for decoding.
We take Transformer as the testbed and introduce a
layer of stochastic gates in-between the encoder and the decoder.
The gates are regularized using the expected value of the sparsity-inducing $L_0$ penalty,
resulting in completely masking-out a subset of encoder outputs.
In other words, via joint training, the \lzdrop{} layer forces Transformer to route information through a subset of its encoder states.
We investigate the effects of this sparsification on two machine translation and two summarization tasks.
Experiments show that, depending on the task, around 40--70\% of source encodings can be pruned without significantly compromising  quality.
The decrease of the output length endows \lzdrop{} with the potential of improving decoding efficiency, where it yields a speedup of up to 1.65$\times$ on document summarization tasks against the standard Transformer.
We analyze the \lzdrop{} behaviour and observe
that it exhibits systematic preferences for pruning certain word types, e.g., 
function words and punctuation get pruned most.
Inspired by these observations, we  explore the feasibility of
specifying rule-based patterns that mask out encoder outputs based on information such as part-of-speech tags, word frequency and word position.\footnote{Source code is available at \url{https://github.com/bzhangGo/zero}.}
\end{abstract}

\section{Introduction}

\begin{figure}[t]
  \centering
    \includegraphics[scale=0.45]{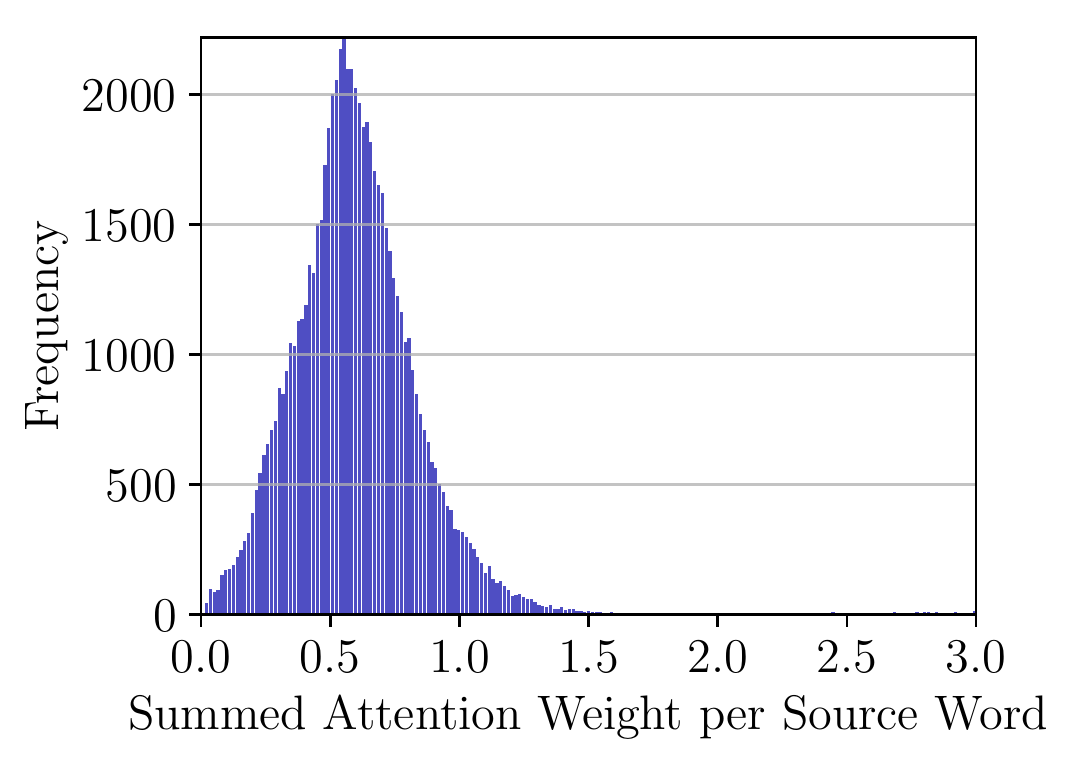}
  \caption{\label{fig_att_hist} Distribution of the summed attention weight per source word estimated on the English-German WMT14 test set. For each (source sentence, translation) pair, we extract the attention matrices from all encoder-decoder attention sublayers in Transformer and average them over different (8) heads and (6) layers. The attention value for each source word is summed over all target words in the translation. Higher attention weights suggest larger impacts on translation. Around 49.7\% source words get attention weights of less than 0.6, compared to the mean value of 1.03. 
  }
\end{figure}

Neural sequence-to-sequence (Seq2Seq) models have dominated various text generation tasks, including machine translation~\cite{NIPS2017_7181} and abstractive document summarization~\cite{gehrmann2018bottom,liu-lapata-2019-hierarchical}. 
These models generally follow the encoder-decoder paradigm, where
the encoder interprets source context and converts source words into vector representations
such that the decoder has sufficient information to predict the target sequence.
Early Seq2Seq models~\cite{sutskever2014sequence,cho2014learning} provided only
the last and/or first encoder states to the decoder.
In contrast, 
modern approaches rely on the attention mechanism~\cite{DBLP:journals/corr/BahdanauCB14}
and implicitly make an 
assumption that information from 
\textit{all} encoder outputs should flow to the decoder.
However, this assumption neglects the fact  that a large portion of source words in machine translation receives just minor attention as shown in Figure \ref{fig_att_hist}, let alone in summarization where the input contains redundant expressions and large parts of text are not relevant to any plausible summary. Moreover, information content varies
across words, for example,  it  is negatively correlated with event frequency~\cite{Shannon:2001:MTC:584091.584093,Zipf49}.\footnote{We interchangeably use \textit{source representation}, \textit{encoder output} and \textit{source encoding} unless otherwise specified.}

\begin{figure}[t]
  \centering
    \includegraphics[scale=0.39]{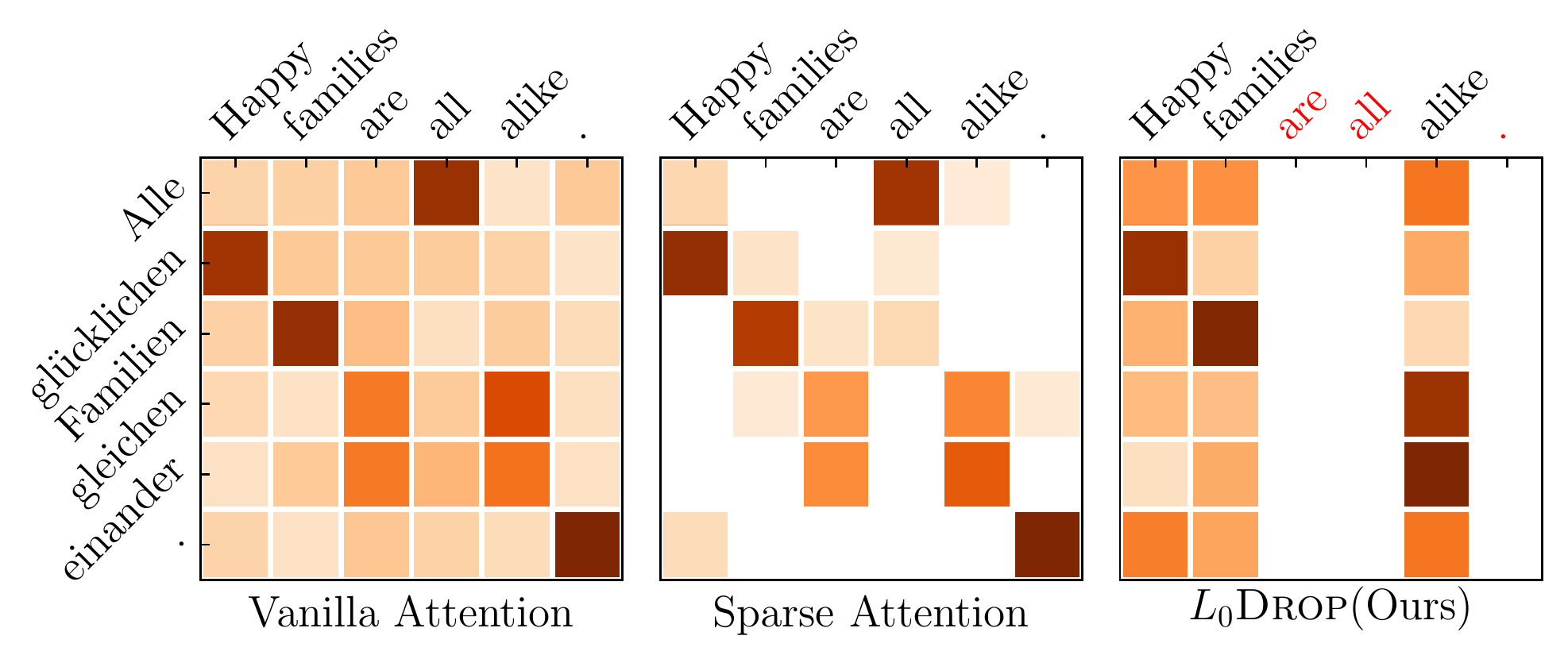}
  \caption{\label{fig_cmp_atts} Encoder-decoder attention distribution of target words (y-axis) over source words (x-axis) for the vanilla attention~\cite{NIPS2017_7181}, the sparse attention~\cite{correia-etal-2019-adaptively} and our model. Darker color indicates larger attention weight, and the white blocks denote an attention weight of 0. The source words whose encoding is pruned by \lzdrop{} (receiving zero weight) are highlighted in red.}
\end{figure}

In this work, we hypothesize that encoder outputs are compressible and we can force Seq2Seq model to route information through their subset.
Figure \ref{fig_cmp_atts} illustrates our intuition as well as the difference with existing work~\cite{NIPS2017_7181,correia-etal-2019-adaptively}. Instead of dynamically sparsifying attention weights for individual decoder steps~\cite{correia-etal-2019-adaptively}, we aim at detecting uninformative source encodings and dropping them to shorten the encoding sequence before generation.
To this end, we build on recent work on sparsifying  weights~\cite{louizos2017learning} and activations~\cite{bastings-etal-2019-interpretable} of neural networks.
Specifically, we insert a differentiable neural sparsity layer (\lzdrop) in-between the encoder and the decoder. 
The layer can be regarded as providing a multiplicative scalar gate for every
encoder output. The gate is a random variable and, unlike standard attention,
can be exactly zero, effectively masking out the corresponding source encodings.
The sparsity is promoted by introducing an extra term to the learning
objective, i.e. an expected value of the sparsity-inducing $L_0$ penalty.
By varying the coefficient for the regularizer, we can obtain different levels of sparsity.
Importantly, the objective remains fully end-to-end differentiable.

Given an encoding sequence of length $N$, the vanilla attention model attends to it recurrently for $M$ steps at the decoding phase, leading to a computational complexity of $\mathcal{O}(NM)$ ($N=6$, $M=6$ in Figure \ref{fig_cmp_atts}). This could be costly if $N$ or $M$ is very large. With the induced sparse structure by \lzdrop{}, we introduce a specialized decoding algorithm which lowers this complexity to $\mathcal{O}(N^\prime M)$ ($N^\prime \leq N$, and $N^\prime=3$ in Figure \ref{fig_cmp_atts}). As a result, \lzdrop{} provides a chance to improve decoding efficiency by reducing the encodings' length, especially for long inputs.

We apply \lzdrop{} to Transformer~\cite{NIPS2017_7181}, the state-of-the-art Seq2Seq model. We conduct extensive experiments on WMT translation tasks with two language pairs and document summarization tasks covering single document and multiple documents settings. We analyze how pruning source encodings impacts the generation quality and which word types get pruned. We also explore rule-based sparsity patterns inspired by the analysis of \lzdrop{}, such as deterministically filtering out the encodings of words with specific POS tags, high-frequency words or simply attending to every other word in the sequence.

Our main findings are summarized as follows:
\begin{itemize}
    \item We confirm that the encoder outputs can be compressed, around 40--70\% of them can be dropped without large effects on the generation quality.
    \item The resulting sparsity level differs across word types, the encodings corresponding to function words (such as determiners, prepositions) are more frequently pruned than those of content words (e.g., verbs and nouns).
    \item \lzdrop{} can improve decoding efficiency particularly for lengthy source inputs. We achieve a decoding speedup of up to 1.65$\times$ on document summarization tasks.
    \item Filtering out source encodings with rule-based sparse patterns is feasible, and confirms information-theoretic expectations, although rule-based patterns do not generalize well across tasks.
\end{itemize}

\section{Related Work}

Approaches to compression in Seq2Seq models fall into the category of model parameter compression~\cite{see-etal-2016-compression}, sequential knowledge distillation~\cite{kim-rush-2016-sequence} or sparse attention induction that ranges from modeling hard attention~\cite{wu-etal-2018-hard} to developing differentiable sparse softmax functions or regularizing attention weights for sparsity~\cite{NIPS2017_6926,correia-etal-2019-adaptively,cui-etal-2019-fine,8550728}.
Unfortunately, the success of all these studies builds upon the access to all source encodings in training and decoding.  
Learning which encoder outputs to prune in Seq2Seq models, to the best of our knowledge, has never been investigated before.
\citet{sukhbaatar-etal-2019-adaptive} learn attention spans in self-attention and
discard information from states outside of the span; this method is not directly applicable to encoder-decoder attention.

We use the differentiable ${L}_0$-relaxation which was first introduced by \citet{louizos2017learning} in the context of pruning individual neural network parameters.
It was previously used to prune heads in multi-head attention~\cite{voita-etal-2019-analyzing}. 
Our work is more similar in spirit to \citet{bastings-etal-2019-interpretable} where they used
the ${L}_0$ relaxations to construct interpretable classifiers, i.e. models that can reveal
which words they rely on when predicting a class. In their
approach, the information from dropped words is lost 
rather than rerouted into the states of retained words, as desirably
for interpretability but problematic in the
text generation set-up.

The number of the source encodings selected by \lzdrop{} is sentence-dependent, which differs from the linear-time model of \citet{wang-2019-towards}, although both can accelerate decoding. Our study of rule-based sparsity patterns is in line with the sparse Transformer~\cite{child2019generating} though we also explore the use of external linguistic information (POS tag) in our sparsification rules, and focus on encoder outputs instead of self-attention.

\section{Background: Transformer}

We take Transformer~\cite{NIPS2017_7181} as our testbed. Transformer uses the dot-product attention network as its backbone to handle intra- and inter-sequence dependencies:
\begin{equation}\label{eq_attention}
    \begin{split}
       \textsc{Att}(\mathbf{H}, \mathbf{M}) = \mathbf{A} \mathbf{V} = \textsc{Sm}\left(\frac{\mathbf{Q}\mathbf{K}^T}{\sqrt{d}}\right) \mathbf{V},
    \end{split}
\end{equation}
where $\mathbf{Q}, \mathbf{K}, \mathbf{V} = \mathbf{H}\mathbf{W}_q, \mathbf{M}\mathbf{W}_k, \mathbf{M}\mathbf{W}_v$. The input $\mathbf{H}\in \mathbb{R}^{J\times d}$ of length $J$ queries and summarizes task-relevant clues from the memory $\mathbf{M} \in \mathbb{R}^{I\times d}$ of length $I$ based on their dot-product semantic matching $\mathbf{A} \in \mathbb{R}^{J \times I}$. $\textsc{Sm}$ denotes the softmax function, $d$ is the model dimension, and $\mathbf{W}_q, $ $\mathbf{W}_k, \mathbf{W}_v \in \mathbb{R}^{d\times d}$ are trainable model parameters. \citet{NIPS2017_7181} also extend this mechanism to multi-head attention.

Given a source sequence $X=\left(x_1, x_2, \ldots, x_N\right)$, Transformer maps it to the  target sequence $Y=\left(y_1, y_2, \ldots, y_M\right)$ following the encoder-decoder paradigm~\cite{DBLP:journals/corr/BahdanauCB14}:\footnote{Each sublayer ($\textsc{Att}$/$\overline{\textsc{Att}}$/$\textsc{Ffn}$) in the encoder and decoder is wrapped with residual connection~\cite{DBLP:journals/corr/HeZRS15} followed by layer normalization~\cite{lei2016layer}, which are dropped in Eq. (\ref{eq_encoder}) and (\ref{eq_decoder}) for clarity.}
\begin{align}
% \begin{split}
    \mathbf{X}^L & = \text{Encoder}\left(\mathbf{X}^0\right)  \label{eq_encoder} \\
                 & \stackrel{L}{\underset{l=1}{:=}} \textsc{Ffn}\left(\textsc{Att}(\mathbf{X}^{l-1}, \mathbf{X}^{l-1})\right), \nonumber
% \end{split}
\end{align}
\begin{align}
% \\
% \begin{split}
    \mathbf{Y}^L & = \text{Decoder}\left(\mathbf{Y}^0, \mathbf{X}^L\right) \label{eq_decoder} \\
                 & \stackrel{L}{\underset{l=1}{:=}} \textsc{Ffn}\left(\textsc{Att}\left(\overline{\textsc{Att}}(\mathbf{Y}^{l-1}, \mathbf{Y}^{l-1}), \mathbf{X}^L\right)\right),\nonumber
% \end{split}
\end{align}
where $\mathbf{X}^0 \in \mathbb{R}^{N\times d}$ and $\mathbf{Y}^0 \in \mathbb{R}^{M\times d}$ stand for the source and the shifted target sequence embedding, respectively, enriched with positional encoding~\cite{NIPS2017_7181}. $\textsc{Ffn}(\cdot)$ is a point-wise feed-forward network. $\overline{\textsc{Att}}(\cdot, \cdot)$ in the decoder denotes masked $\textsc{Att}(\cdot, \cdot)$ which prevents access to future target words. Both the encoder and the decoder involve a stack of $L=6$ identical layers, with the encoder output $\mathbf{X}^L$ fed to the decoder via an encoder-decoder attention sublayer, i.e. the $\textsc{Att}(\cdot, \cdot)$ in Eq. (\ref{eq_decoder}).

Based on the decoder output $\mathbf{Y}^L$, Transformer performs the next-word prediction and adopts the maximum likelihood loss for training.
% \begin{equation}\label{eq_objective_mle}
%     \begin{split}
%         % \mathcal{L}_{\textsc{Mle}}\left(X, Y\right) & = \sum_{i=1}^m \log p\left(y_i|X, Y_{<i}\right) \\
%         %             & = \sum_{i=1}^m \log \textsc{Sm}\left(\mathbf{W}_o \mathbf{Y}^L_{i}\right),
%         % \mathcal{L}_{\textsc{Mle}}\left(X, Y\right) = - \sum_{i=1}^M \log \textsc{Sm}\left(\mathbf{W}_o \mathbf{y}^L_{i}\right),
%         \mathcal{L}_{\textsc{Mt}}\left(X, Y\right) & = - \log p(Y|X).
%     \end{split}
% \end{equation}
% where $\mathbf{y}^L_i$ denotes the $i$-th decoder representation of $\mathbf{Y}^L$, and $\mathbf{W}_o$ is a model parameter that projects $\mathbf{y}^L_i$ onto the target vocabulary space.

\section{Neural Sparsity Layer: \lzdrop{}}

In this section, we introduce a neural sparsity layer (\lzdrop{}), which 
we use to prune encoder outputs. At inference time, only retained encoder outputs will be used as input to the decoder. 

\subsection{Training with \lzdrop{}}

Intuitively, \lzdrop{} assigns each encoder output $\mathbf{x}_i^L$ a gate $g_i \in [0, 1]$ ($i \in \{1,\ldots,N\}$) 
\begin{align}
    \text{\lzdrop{}}(\mathbf{x}_i^L) & = g_i \mathbf{x}_i^L \label{eq_l0drop}, 
\end{align}
and prunes encodings by closing their gates, i.e. $g_i = 0$, relying on adding a differentiable sparsity-inducing penalty to the objective.  

\begin{figure}[t]
  \centering
    \includegraphics[scale=0.50]{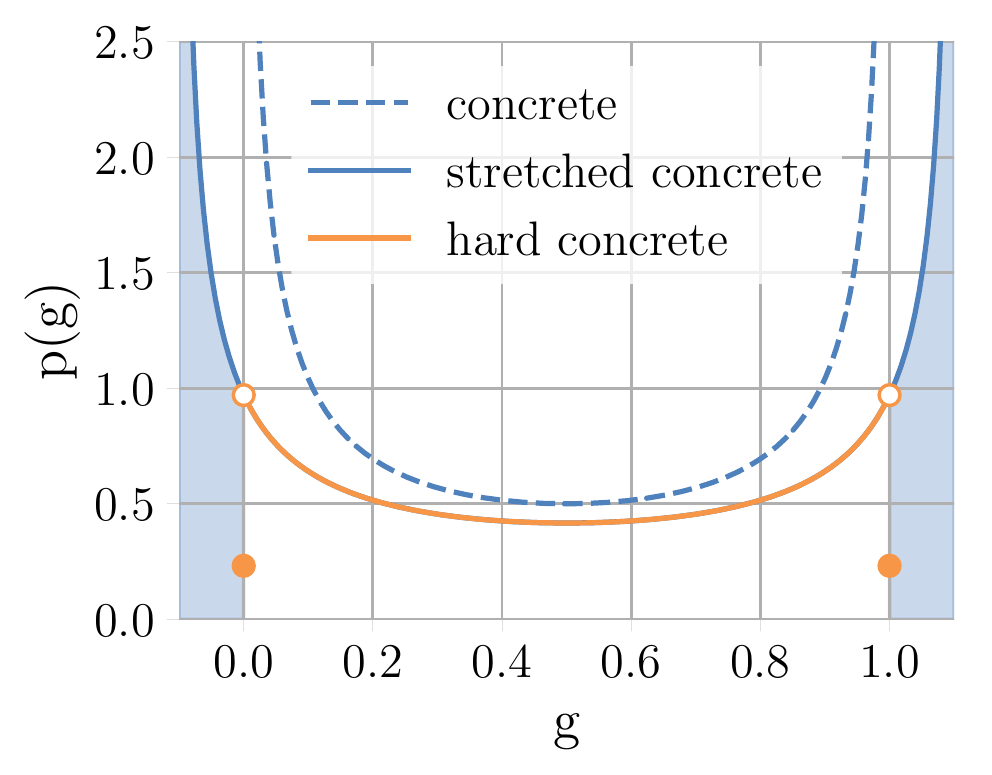}
  \caption{\label{fig_concrete} Hard concrete distribution (orange curve): samples are first stretched from the binary concrete distribution (dashed blue curve) to a stretched distribution (solid blue curve), and then rectified to collapse the probability mass of the shadow areas into \{0\} and \{1\} (solid orange points).}
\end{figure}

More formally, to achieve sparsity, each gate is assumed to be a random variable and its value is
drawn from the HardConrete distribution:
\begin{align}
    g_i  \thicksim \text{HardConcrete}(\alpha_i, \beta, \epsilon),
\end{align}
where $\alpha_i$, $\beta$ and $\epsilon$ are shape parameters of the distribution. 
HardConcrete \cite{louizos2017learning} is a parameterized family of mixed discrete-continuous distributions over the closed interval $[0,1]$. These distributions have point mass at $0$ and $1$ and continuous
density in-between, i.e. in $(0,1)$, as shown in Figure \ref{fig_concrete}.
Thus, the gates will have a non-zero probability of being exactly $0$, corresponding to masking out the input completely.  

Specifically, the sample from HardConcrete distribution is obtained by
stretching and rectifying samples from 
BinaryConcrete distributions~\cite{maddison2016concrete,jang2016categorical}:
\begin{align}
    s_i & \thicksim \text{BinaryConcrete}(\alpha_i, \beta) \label{eq_binary_concrete}\\
    \bar{s}_i & = s_i \left(1 + 2  \epsilon \right) - \epsilon, \label{eq_stretch} \\
    g_i & = \min \left(1, \max \left(0, \bar{s}_i\right)\right). \label{eq_rectify}
\end{align}
In the above expression, we first obtain a sample from the BinaryConcrete distribution (Eq. (\ref{eq_binary_concrete})), then stretch it from  $(0, 1)$ to $(-\epsilon, 1 + \epsilon)$ (Eq. (\ref{eq_stretch}), $\epsilon >0$), and finally rectify with a hard sigmoid  to the closed interval $[0,1]$ (Eq. (\ref{eq_rectify})). 

Note that the  probability of $g_i$ being exactly $0$ ($p(g_i = 0 | \alpha_i, \beta, \epsilon)$)
equals the probability of $\bar{s}_i$ hitting $(-\epsilon,0)$ and is available in a closed
form~\cite{louizos2017learning}:
\begin{align}
\nonumber
p(g_i = 0 | \alpha_i, \beta, \epsilon) = \sigma (  \beta \log \frac{\epsilon}{1 +  \epsilon} - \log \alpha_i),
\end{align}
where $\sigma(\cdot)$ denotes the sigmoid function.
The  parameter $\alpha_i$ (i.e. the location parameter of BinaryConcrete)  
is predicted relying on the encoder output $\mathbf{x}_i$:
\begin{align}
\log \alpha_i = \mathbf{x}_i^L \mathbf{w}^T,
\end{align}
where $\mathbf{w} \in \mathbb{R}^{d}$ is a learned parameter vector;
 the temperature $\beta$ and the stretch degree $\epsilon$ are treated as hyperparameters.
By adjusting $\alpha_i$ the model can change the shape of the HardConcrete distribution,
and dynamically decide which outputs to pass to the decoder and which to prune.

Note that the sum  
\begin{align}
\label{eq_sparsity_penalty}
\mathcal{L}_0(X) = \sum_{i=1}^{N}{1 - p(g_i = 0 | \alpha_i, \beta, \epsilon )},
\end{align} 
yields the expected number of open gates, or, equivalent, the expected $L_0$
loss on gate vector $(g_1,\ldots,g_N)$. Minimizing the loss encourages the model to prune encoder outputs.

Once \lzdrop{} is integrated as a new layer into Transformer, the decoder, previously defined in Eq. (\ref{eq_decoder}), becomes:
\begin{equation}\label{eq_decoder_ldrop}
    \mathbf{Y}^L = \text{Decoder}(\mathbf{Y}^0, \text{\lzdrop{}}(\mathbf{X}^L)).
\end{equation}
Other components in Transformer are kept intact, except for using a modified training objective $\mathcal{L}(X, Y)$:
\begin{align}
   & \mathcal{L}_{\textsc{Mle}}(X, Y)  + \lambda \mathcal{L}_0(X) \nonumber \\
   = &  - \log \mathbb{E}_{\mathbf{g}\sim p(\mathbf{g}|\phi)} \left[ p(Y, \mathbf{g}|X) \right]  + \lambda \mathcal{L}_0(X) \nonumber  \\
   \le & ~ \mathbb{E}_{\mathbf{g}\sim p(\mathbf{g}|\phi)} \left[ - \log p(Y, \mathbf{g}|X) \right]  + \lambda \mathcal{L}_0(X) \nonumber  \\
   = & \mathcal{L}(X, Y) 
\label{eq_objective}
\end{align}
where $\phi$ is short for $(\mathbf{\alpha}, \beta, \epsilon)$, $\lambda \in \mathbb{R}^+$ is a hyperparameter defining the level of sparsity. The bound is derived by applying  Jensen's inequality.

Importantly, the objective remains fully differentiable as
we can rely on  the reparameterization technique~\cite{kingma2013auto} to sample $\tilde{\mathbf{g}}$ for  
computing unbiased estimates of the gradients.
Adding \lzdrop{} and the regularizer introduces only a negligible computational overhead to training compared to the original Transformer.

\subsection{Decoding with \lzdrop{}}

At test time we do not sample gate values but 
estimate their expected value $g_i$  as follows~\cite{louizos2017learning}:
\begin{equation}\label{eq_z_mean}
    \hat{g}_i = \min (1, \max (0, \sigma(\log \alpha_i)(1 + 2\epsilon) - \epsilon)),
\end{equation}
which often turns out to be exactly either 0 or 1, albeit being in-between in some cases. Encodings of non-zero $\hat{g}_i$ are preserved and simply weighted by the gate. 

\begin{algorithm}[t]
    \caption{\label{alg_decoding} Algorithm for the encoder-decoder attention with \lzdrop{} at decoding}
    \begin{algorithmic}[1]
        \REQUIRE Source encodings, $\mathbf{X}^L \in \mathbb{R}^{N\times d}$; \\
        		\quad~ Gates, $\hat{\mathbf{g}} \in \mathbb{R}^N$; \\
        		\quad~ Query state, $\mathbf{y}_j^l \in \mathbb{R}^d$; \\
        \ENSURE Attention vector for the query

        $\triangleright$ {\em step 1: reorganize source-side inputs}\\
        \STATE $I$ $\leftarrow$ $\{i|\hat{g}_i \neq 0\}$ $\quad$ $\triangleright~~ N^\prime$ $\leftarrow$ $|I|$ \label{alg_decoding_step1_start}
        \STATE $\hat{\mathbf{g}}^\prime$ $\in$ $\mathbb{R}^{N^\prime}$, ${\mathbf{X}^\prime}^L$ $\in$ $\mathbb{R}^{N^\prime \label{algo_decoding_step1_middle} \times d}$ $\leftarrow$ $\hat{\mathbf{g}}[I]$, $\mathbf{X}^L[I]$
        \STATE $c$ $\leftarrow$ $N - N^\prime$
        \STATE $\bar{\mathbf{X}}^L \leftarrow [\mathbf{0}\in \mathbb{R}^d, {\mathbf{X}^{\prime}}^L \odot \hat{\mathbf{g}}^\prime]$, ${\mathbf{c}} \leftarrow [c, \mathbf{1} \in \mathbb{R}^{N^\prime}]$ \label{alg_decoding_step1_end}
        $\triangleright$ {\em step 2: attention with counts} \\
        \STATE $\mathbf{q}, \mathbf{K}, \mathbf{V} \leftarrow \mathbf{y}_j^l \mathbf{W}_q, \bar{\mathbf{X}}^L \mathbf{W}_k, \bar{\mathbf{X}}^L \mathbf{W}_v$ \label{alg_decoding_step2_start}
        \STATE $\mathbf{e} \in \mathbb{R}^{N^\prime + 1} \leftarrow \mathbf{q} \mathbf{K}^T / \sqrt{d}$
        \\
        $\triangleright$ {\em perform softmax with counts}
        \STATE $\mathbf{a} \leftarrow \mathbf{c} \odot \exp(\mathbf{e}) / \sum_t\left(c_t \exp(e_t)\right)$
        \STATE $\mathbf{v} \in \mathbb{R}^d \leftarrow \mathbf{a} \mathbf{V}$ \label{alg_decoding_step2_end}

        \RETURN $\mathbf{v}$
    \end{algorithmic}
\end{algorithm}

To leverage the induced sparse structure, we revise the decoding procedure as in Algorithm \ref{alg_decoding}.
The notation $[\cdot, \cdot]$ refers to row-wise concatenation, $[I]$ stands for extracting elements with the indices $I$, $\odot$ is element-wise multiplication, and $\mathbf{1} \in \mathbb{R}^{N^\prime}$ indicates a vector of ones of length $N^\prime$. 
We first reorganize the gates $\hat{\mathbf{g}}\in \mathbb{R}^N$ and the source encodings $\mathbf{X}^L \in \mathbb{R}^{N\times d}$ by eschewing the entries corresponding to closed gates ($\hat{g}_i=0$, line \ref{alg_decoding_step1_start}-\ref{algo_decoding_step1_middle}). We augment the compressed sequence ${\mathbf{X}^\prime}^L\in \mathbb{R}^{N^\prime \times d}$ with a dummy zero encoding vector $\mathbf{0} \in \mathbb{R}^d$ to represent all pruned encodings, and record their count into a counting vector $\mathbf{c} \in \mathbb{R}^{N^\prime + 1}$ (line \ref{alg_decoding_step1_end}).\footnote{Note that $N^\prime \leq N$. \lzdrop{} could increase the sequence length if no source encoding is pruned, which is not observed in our experiments.} We then modify the attention process to enable the inclusion of this counting information (line \ref{alg_decoding_step2_start}-\ref{alg_decoding_step2_end}) for correctly estimating the attention weights. Note that the shortened source sequence $\bar{\mathbf{X}}^L$ is reused across decoder layers and steps. \lzdrop{} changes the dependency of the encoder-decoder attention on source sequence from $\mathcal{O}(NM)$ to $\mathcal{O}(N^\prime M)$, and allows for efficiency gains even with moderate sparsity, especially for large $L$, $N$ and $M$.
\begin{figure*}[t]
    \centering
    \subfigure[\label{fig_sparse_mt_ende}WMT14 En-De]{
        \begin{minipage}[t]{0.32\linewidth}
        \centering
        \includegraphics[scale=0.42]{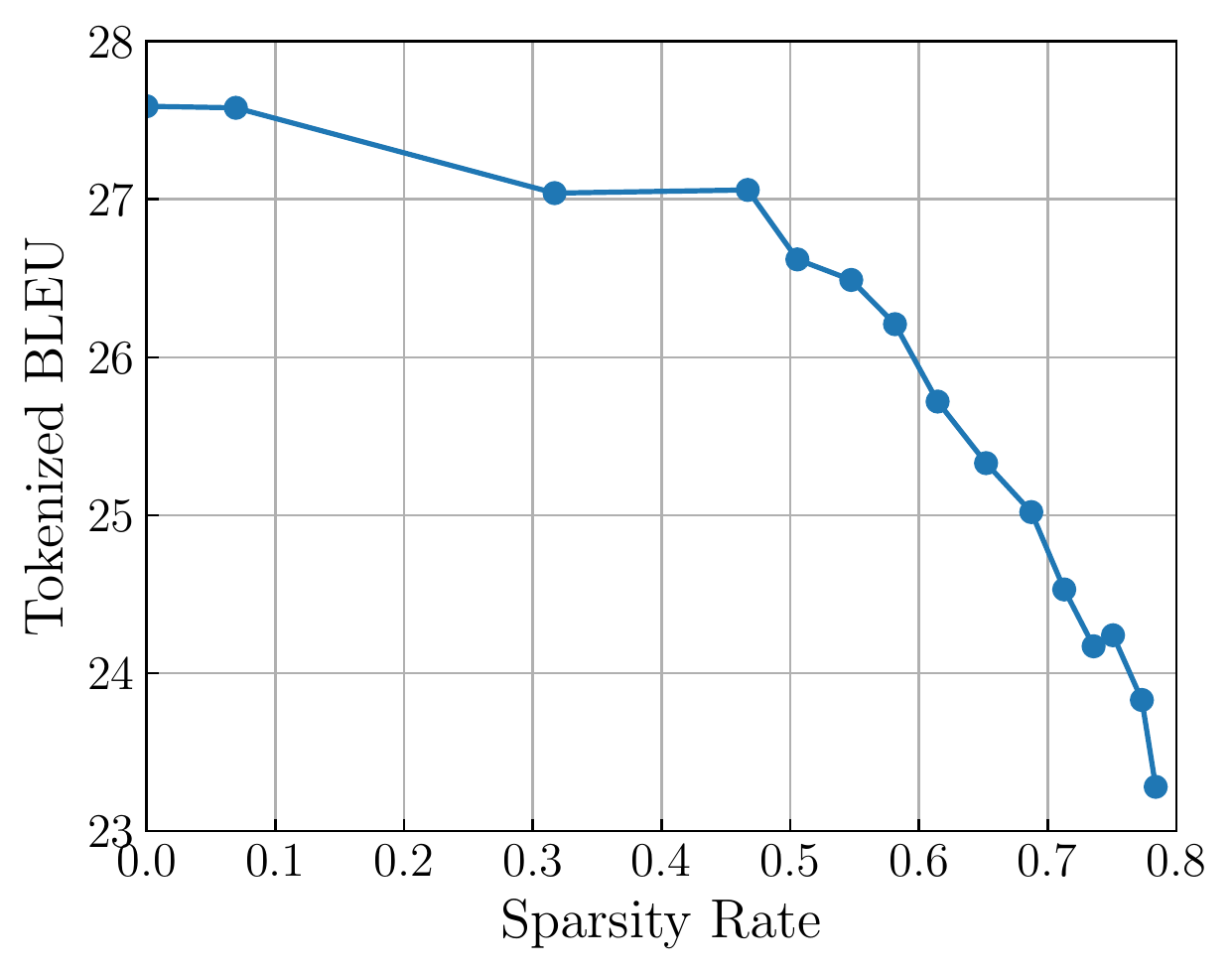}
        \end{minipage}%
    }~%
    \subfigure[\label{fig_sparse_mt_zhen}WMT18 Zh-En]{
        \begin{minipage}[t]{0.32\linewidth}
        \centering
        \includegraphics[scale=0.42]{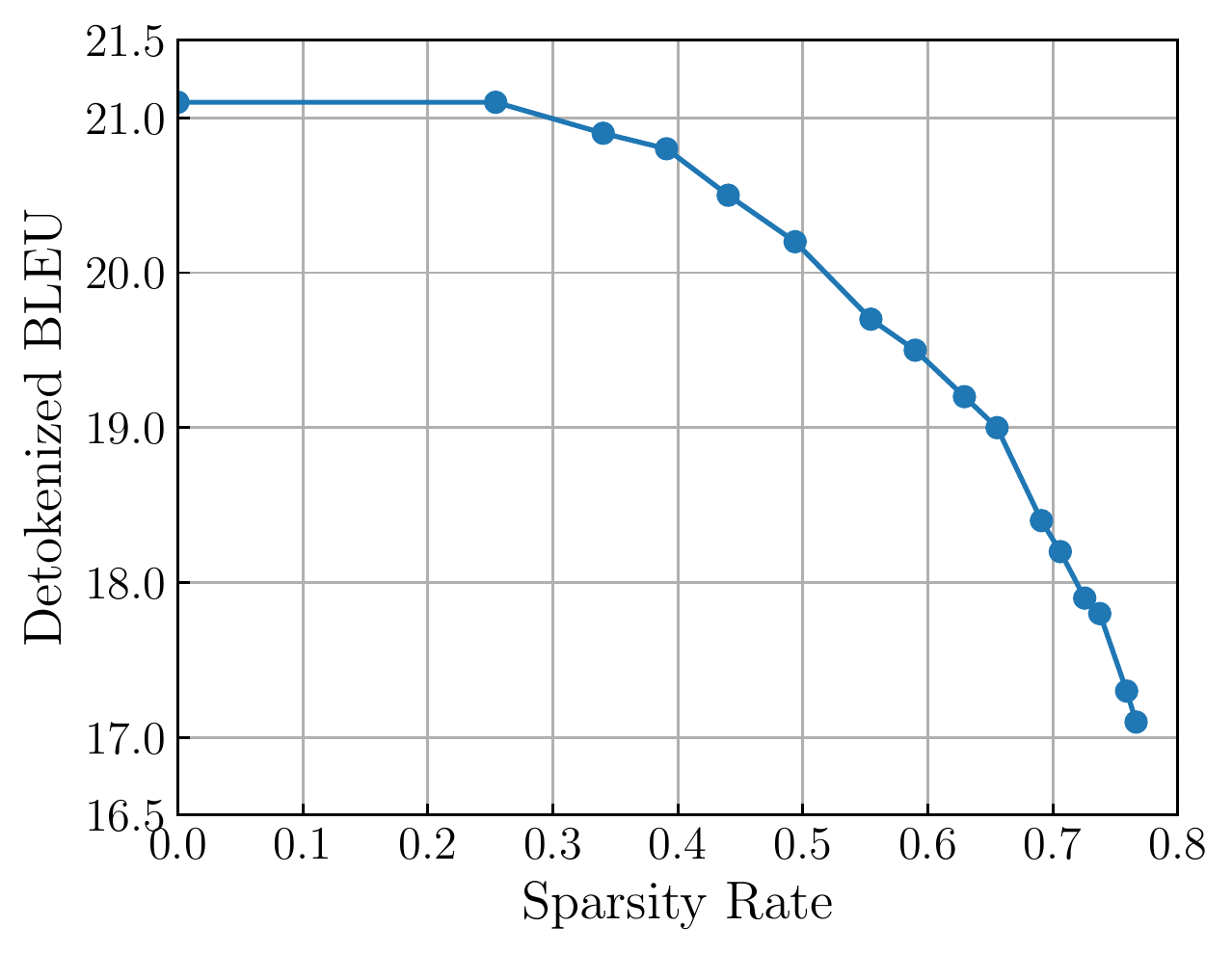}
        \end{minipage}%
    }~%
    \subfigure[\label{fig_sparse_sum_cnn}CNN/Daily Mail]{
        \begin{minipage}[t]{0.32\linewidth}
        \centering
        \includegraphics[scale=0.42]{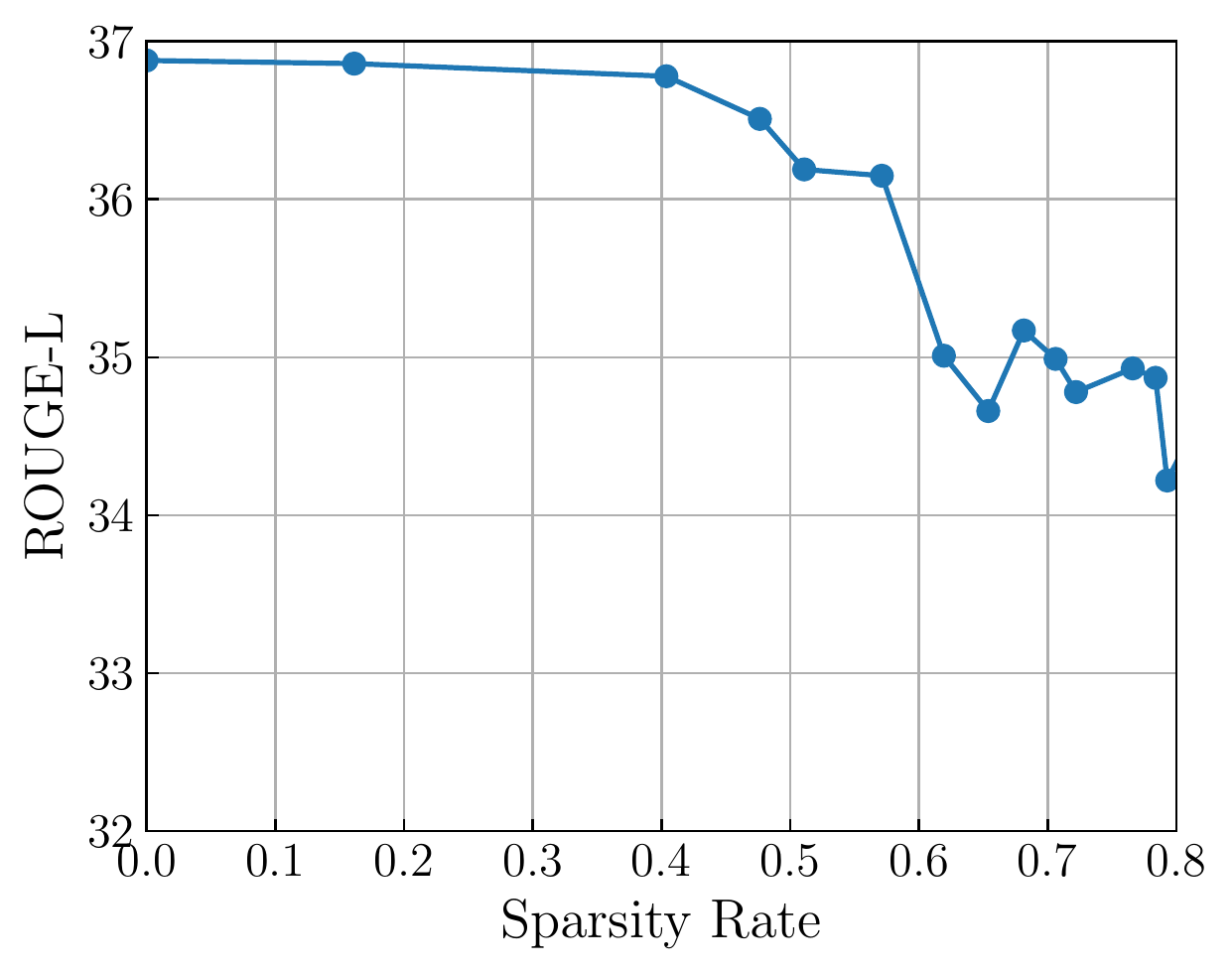}
        \end{minipage}%
    }%
    \caption{\label{fig_sparse}Generation quality (BLEU and ROUGE-L, evaluated on test set) as a function of sparsity rate for WMT14 En-De, WMT18 Zh-En and CNN/Daily Mail. Pruning about 40\% source encodings results in marginal performance loss on all tasks.}
\end{figure*}
\section{Experimental Setup}

% We examine the effectiveness of \lzdrop{} on WMT translation tasks (English-German and Chinese-English) and document summarization tasks (single- and multi-document summarization).

\paragraph{Machine Translation} We train translation models on the WMT14 English-German translation task (En-De)~\cite{bojar-EtAl:2014:W14-33} and the WMT18 Chinese-English translation task (Zh-En)~\cite{bojar-etal-2018-findings}. We use newstest2013 as the validation set for WMT14 En-De and newstest2017 for WMT18 Zh-En. We evaluate the translation quality with BLEU metric~\cite{papineni-etal-2002-bleu}, and report tokenized BLEU on newstest2014 for WMT14 En-De and detokenized BLEU on newstest2018 for WMT18 Zh-En using \textit{sacreBLEU}~\cite{post-2018-call}. We apply the byte pair encoding (BPE) algorithm~\cite{sennrich-etal-2016-neural} with 32K merging operations to handle rare words for both translation tasks. 

\paragraph{Document Summarization} We train abstractive summarization models on the CNN/Daily Mail dataset~\cite{NIPS2015_5945} and the WikiSum dataset~\cite{j.2018generating} for single- and multi-document summarization task, respectively. We use the non-anonymized version of CNN/Daily Mail~\cite{gehrmann2018bottom}. We pre-process this dataset with a BPE vocabulary of 32K and truncate each article to 400 subwords~\cite{gehrmann2018bottom}. 
We use the ranked version of WikiSum~\cite{liu-lapata-2019-hierarchical}, where top-40 paragraphs are extracted for each instance paired with a summary of 121 words on average. We concatenate all these paragraphs into one source sequence following the given ranking order. We employ BPE preprocessing following ~\citet{liu-lapata-2019-hierarchical} and truncate each source sequence to 2048 subwords. We evaluate the summarization quality using the F$_1$ score of ROUGE-L~\cite{lin-2004-rouge}.

\paragraph{Model Settings} We formulate all the above tasks as sequence-to-sequence tasks, and experiment with the base setting of Transformer~\cite{NIPS2017_7181}: $d=512$, the middle layer size of $\textsc{Ffn}(\cdot)$ is 2048, and the number of attention head is 8. Following~\citet{louizos2017learning}, we set $\epsilon=-0.1$, and $\beta=\nicefrac{2}{3}$ for \lzdrop{}. We tune the hyperparameter $\lambda$ for different tasks, as discussed in detail in the following sections. Extra details are provided in Appendix.

\section{Results and Analysis}

\textit{How much can encoder outputs be sparsified?} We answer this question by analyzing the impact of pruning source encodings on the generation quality. We first train a baseline Transformer model, and then finetune this model using \lzdrop{} (Eq. (\ref{eq_objective})) with varied $\lambda$ to explore different levels of sparsity. We sample $\lambda$ with a range of $(0, 1.5]$ and a step size of $0.1$, and finetune WMT14 En-De and WMT18 Zh-En models for extra 50K steps, and CNN/Daily Mail for extra 20K steps. We use the \textit{sparsity rate} to measure the sparsity; we define it as the ratio of the pruned source encoding number $\#(\hat{g}_i=0)$ to the total number of source words.

Figure \ref{fig_sparse} shows the results. The generation quality exhibits a negative correlation with the sparsity rate across different tasks, reflecting the usefulness of encoder outputs for generation. However, the fact that we can remove about 40\% source encodings without largely degrading the generation performance (-0.5 BLEU and -0.1 ROUGE-L) supports our hypothesis that we can force Seq2Seq model to route information through a subset of its source encodings. 
We also observe that the compressibility seems relatively language independent (the curves of WMT14 En-De \ref{fig_sparse_mt_ende} and WMT18 Zh-En \ref{fig_sparse_mt_zhen} are similar) but clearly task dependent. Compared to translation tasks, the summarization task is less sensitive to the pruning of source encodings (-1.89 ROUGE-L versus -3.0 BLEU at a sparsity rate of $\sim$70\%). We ascribe this to the property of summarization where the summary only reflects a part of the input document, rather than the entire document.

\begin{figure}[t]
  \centering
    \includegraphics[scale=0.45]{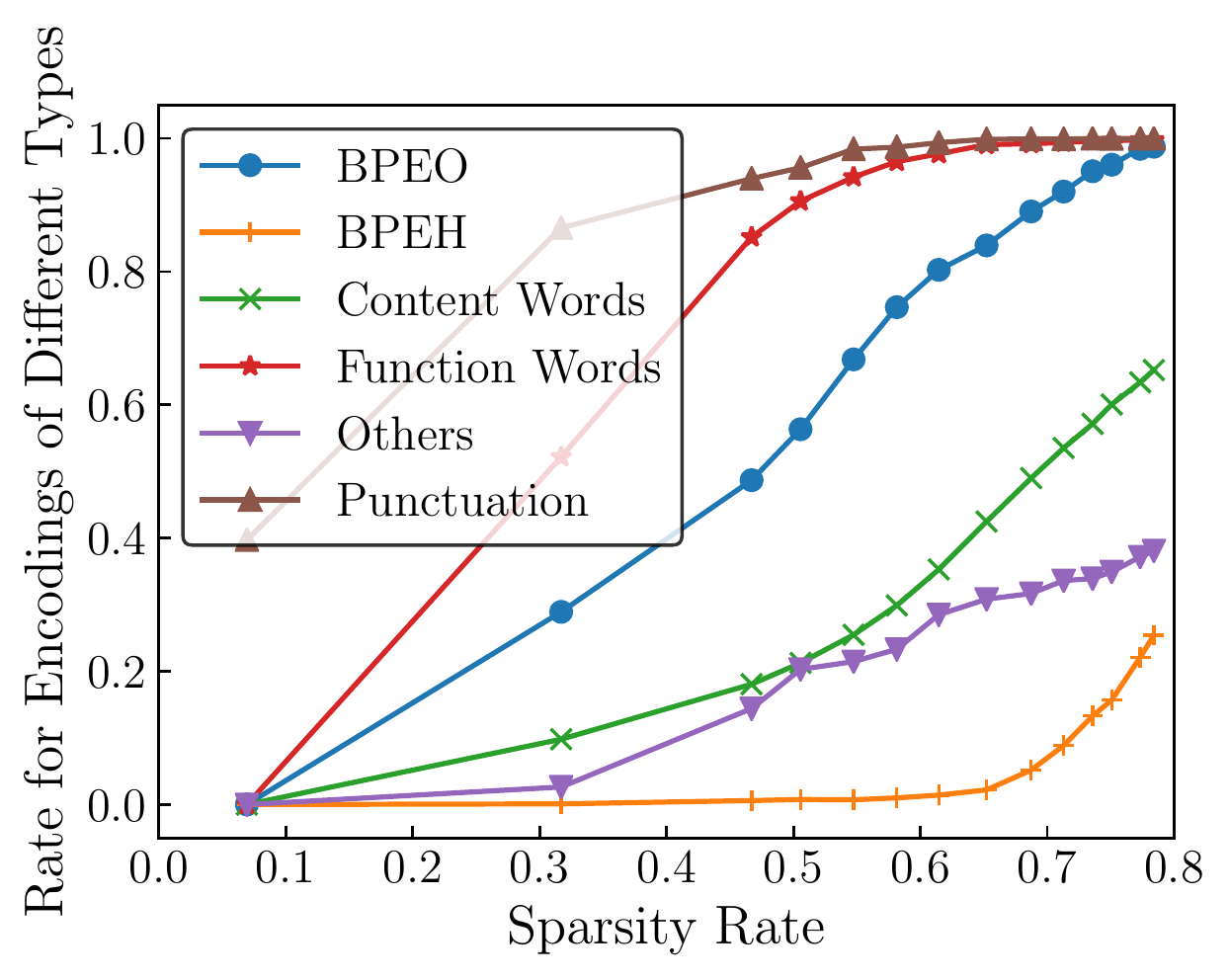}
  \caption{\label{fig_sparse_pos_analysis} Curves for sparsity rate of different types of encoding on the WMT14 En-De test set. $x$-axis denotes the overall sparsity rate. The encoding of content words and BPEH is more valuable for generation, compared to that of function words and punctuation.}
\end{figure}
Note that the pretraining-then-finetuning schema is mainly used for saving training efforts. By scheduling $\lambda$ linearly with training steps, we can train models with \lzdrop{} (Eq. (\ref{eq_objective})) from scratch, and obtain a BLEU score of 27.03 ($\lambda=0.2$, warm-up step of 200K) on WMT14 En-De, comparable to using finetuning (27.04).

\textit{What types of source encoding are required for generation?}
Our goal here it to understand encodings of which types of tokens are retained. For each source encoding, we regard the POS of its corresponding word as its type. We take WMT14 En-De as our benchmark, where we annotate POS for source sentences in the test set using the Stanford POS tagger~\cite{toutanova-etal-2003-feature}. We handle subwords separately by labeling its first piece as \textit{BPEH} while the others as \textit{BPEO}, regardless of the POS of its unsegmented form. We group different POS tags into 6 categories for the sake of analysis: BPEH, BPEO, function words, content words, punctuation and the rest.\footnote{Function words include \textit{CC, IN, RP, TO, UH, DT} and \textit{WP}. Content words include \textit{MD, JJ, NN, RB} and \textit{VB}. Others include other POS tags except for punctuation and BPEO/BPEH, such as \textit{CD, EX, FW} and \textit{SYM}.}

Figure \ref{fig_sparse_pos_analysis} shows 
how the sparsity rate of each encoding type
changes as a function of the overall sparsity rate. We find that \lzdrop{} first choose to eliminate the encoding of punctuation, followed by that of function words. These words often signal structural and grammatical relationships that, while important to build up a representation of the sentence, can be easily compressed. In contrast, pruning content words, which express richer lexical meaning, is more difficult. The sharp increase of content word sparsity after the overall sparsity rate of 0.5 in Figure \ref{fig_sparse_pos_analysis} correlates with a sharp drop in translation quality (see Figure \ref{fig_sparse_mt_ende}). 
We also observe that there is a large difference between BPEO and BPEH, albeit both from the same word. \lzdrop{} favours to prune the encoding of BPEO, indicating that the model learns to use word-initial representations (BPEH) to represent whole words.

\textit{What's the effect of \lzdrop{} on Transformer?} Transformer can lose the access to around 40\% source encodings while largely retaining the same performance. We try to figure out what has changed inside Transformer in order to support \lzdrop{}, and analyze the attention weights (i.e. $\mathbf{A}$ in Eq. (\ref{eq_attention})) of all encoder-decoder attention sublayers and the last encoder self-attention sublayer; these sublayers are directly connected with \lzdrop{} in the computation graph. We experiment on WMT14 En-De.

\begin{figure}[t]
  \centering
    \includegraphics[scale=0.50]{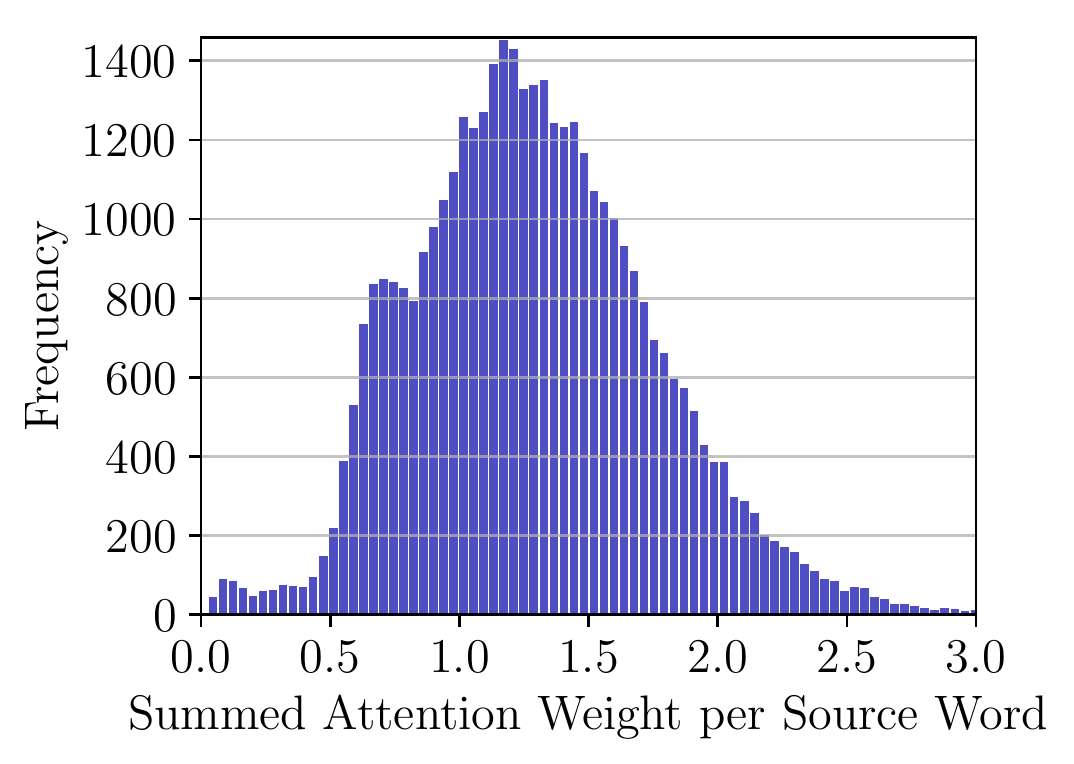}
  \caption{\label{fig_att_hist_l0} Distribution of the summed attention weight per source word on the WMT14 En-De test set for Transformer with \lzdrop{} (sparsity rate 47\%, BLEU 27.06). Only 4.5\% source words get attention weights of less than 0.6.}
\end{figure}

We visualize the distribution of the encoder-decoder attention weight per source word for Transformer with a sparsity rate of 47\% (BLEU 27.06). Compared to the vanilla Transformer (Figure \ref{fig_att_hist}), distributions in Figure \ref{fig_att_hist_l0} show that the average attention weight obtained by each source word has increased (+0.77, 1.03$\rightarrow$1.80), and the proportion of source words receiving attention weights of less than 0.6 is substantially reduced, by a factor of 10 (49.7\%$\rightarrow$4.5\%).
This indicates that \lzdrop{} forces Transformer to distribute its attention more evenly among the retained source encodings.

\begin{figure}[t]
  \centering
    \includegraphics[scale=0.50]{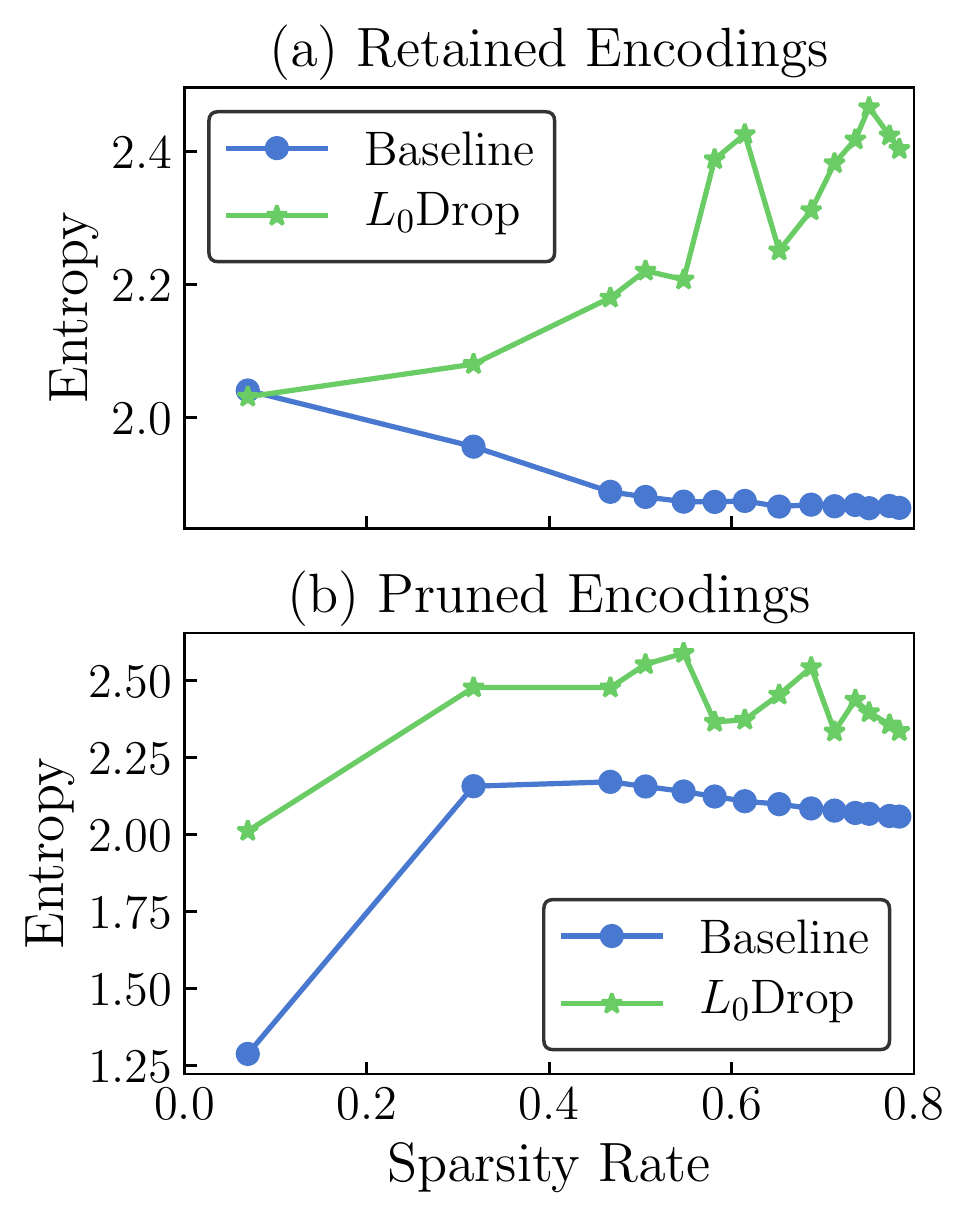}
  \caption{\label{fig_entropy} Entropy of the retained source encodings (top) and the pruned ones (bottom) versus the sparsity rate on the WMT14 En-De test set. We use the sparsity variable $\mathbf{\hat{g}}$ learned by \lzdrop{} to classify the encodings of our baseline Transformer. Higher entropy indicates that the distribution tends to be uniform. With fewer retained encodings, Transformer tends to spread its attention weights to include more source-side information.}
\end{figure}

Apart from the encoder-decoder attention, we also inspect the self-attention in the last encoder layer. We average the self-attention weights over 8 different heads, and compare the attention entropy of the retained source encodings ($\hat{g}_i \neq 0$) and the pruned ones ($\hat{g}_i = 0$). We report average entropy values over the whole test set. Figure \ref{fig_entropy} shows how increasing sparsity affects the entropy. Although \lzdrop{} selects to drop uninformative encodings, the increase in the entropy of the retained encodings (Figure \ref{fig_entropy} (a)), when compared to the baseline, suggests that the encoder actually encodes more context information into these representations, confirming that the model learns to compress context information when sparsity is enforced. Another observation is that the entropy curve of \lzdrop{} for the pruned encodings is in line with that of the baseline, albeit on a larger scale (Figure \ref{fig_entropy} (b)). This signifies that \lzdrop{} adapts Transformer to better coordinate with source context representations, which ensures its effectiveness on generation.

\textit{Can we prune encodings earlier in the encoder?} 
Rather than stacking \lzdrop{} on top of the encoder outputs, we insert \lzdrop{} in-between every adjacent pair of encoder layers. We work on WMT14 En-De and finetune with $\lambda=0.2$. We get a sparsity rate of 0.0\%, 0.0\%, 8.6\%, 8.6\%, 8.7\% and 34.0\% for the first to the last \lzdrop{} layer, respectively, with a BLEU score of 26.74. This result suggests that Transformer does not gain much benefit from pruning encodings  earlier. The model tends to retain encodings at shallow levels (0.0\%/8.6\% $<$ 34.0\%), and loses 0.3 BLEU compared to its \lzdrop{} baseline  ($\lambda=0.2$, sparsity rate $31.7\%$, BLEU 27.04). We believe that the encoder relies on low-level information (including the words) to fully `understand' the sentence, though part of the final encodings is discardable.

\begin{table}[t]
\centering
\small
\begin{tabular}{p{2cm}cccc}
\toprule
Task & \multicolumn{2}{p{2cm}}{Time~Speedup} & Sparsity & Quality \\
\cmidrule(lr){2-3}
\cmidrule(lr){4-5}
\multirow{2}{*}{WMT14 En-De} 
    & 68.89 & 1.00$\times$ & 0.00\% & 27.59 \\
    & 68.38 & 1.01$\times$ & 46.7\% & 27.06 \\
\addlinespace
\multirow{2}{*}{WMT18 Zh-En}
    & 116.3 & 1.00$\times$ & 0.00\% & 21.10 \\
    & 118.3 & 0.98$\times$ & 39.1\% & 20.80 \\
\addlinespace
\multirow{2}{*}{CNN/Daily Mail}
    & 3909 & 1.00$\times$ & 0.00\% & 36.88 \\
    & 3227 & 1.21$\times$ & 47.6\% & 36.51 \\
\addlinespace
\multirow{2}{*}{WikiSum}
    & 70505 & 1.00$\times$ & 0.00\% & 39.20 \\
    & 42669 & 1.65$\times$ & 71.5\% & 38.75 \\
\bottomrule
\end{tabular}
\caption{\label{tb_decoding_speed} Decoding results for different tasks when finetuning with $\lambda=0.3$. ``Time'': the decoding time (in seconds) of the whole test set. ``Sparsity'': the sparsity rate, 0.00\% indicates the Transformer baseline. ``Speedup'': the decoding acceleration over the baseline. ``Quality'': BLEU for WMT tasks and ROUGE-L for summarization tasks. We evaluate the decoding time on GeForce GTX 1080 Ti, with a batch size of 32 for WMT tasks and 10 for summarization tasks.} 
\end{table}

\textit{Can we make the decoding faster with \lzdrop{}?} With appropriate finetuning, \lzdrop{} can shorten the encoding sequence fed to the decoder, reducing the calculation amount of the encoder-decoder attention. However, the encoder-decoder attention corresponds to about \nicefrac{1}{3} of the decoder calculations,\footnote{At decoding, the encoder-decoder attention accounts for about 34.4\% decoder time according to profiling on our implementation.} and Algorithm \ref{alg_decoding} also brings in extra overhead, such as gathering and indexing operations. Thus, a speed-up is not guaranteed, and we report empirical decoding time across different tasks.

Results in Table \ref{tb_decoding_speed} show that \lzdrop{} only marginally improves the decoding speed for machine translation, despite a high sparsity rate of 46.7\% (WMT14 En-De) and 39.1\% (WMT18 Zh-En). By contrast, \lzdrop{} yields a speedup of 1.21$\times$ and 1.65$\times$ on CNN/Daily Mail and WikiSum, respectively. One explanation lies at the significant difference in target sequence length, where the average length per summary is $>$60, compared to $\sim$25 in machine translation. Note that \lzdrop{} achieves a substantially higher sparsity rate of 71.5\% on WikiSum with the same $\lambda=0.3$. This is because the input paragraphs overlap in content; the information about redundant words  does not need to be routed into other encoder states,  making easier to prune them.

\section{Exploring Rule-based Sparse Patterns}

Our analysis shows that the sparsity induced by \lzdrop{} follows certain patterns, with the encodings of `less content-bearing' words pruned first. This suggests that we may be able to define heuristic patterns manually.
In this section, we explore the following three rule-based patterns according to our study on WMT14 En-De:
\begin{description}
    \item[\normalfont \textit{POS Pattern}] This pattern discards the source encodings of those easy-to-prune types, including function words, punctuation, BPEO and \textit{MD, EX}, which account for 46.4\% of the source-side WMT14 En-De training data.
    \item[\normalfont \textit{Freq Pattern}] Inspired by the fact that punctuation and function words are high frequency words, we propose to filter out the source encodings corresponding to top-frequent words with a threshold of 46.3\% (top 100 words). We also include an inverse version, \textit{Inv Freq Pattern}, for comparison, which drops the encodings of most rare words; source words whose frequency ranks lower than 452 are removed, covering $\sim$40.0\% of the source training data.
    \item[\normalfont \textit{Group Pattern}] We explore a position-based pattern that only feeds the encodings at odd positions to the decoder, indicating a sparsity rate of $\sim$50\%. This pattern is partially motivated by~\citet{child2019generating}.
\end{description}
Note that the design of these patterns follows our analysis on \lzdrop{}, where we match the sparsity rate in each pattern to the optimal rate of \lzdrop{} on WMT14 En-De. We examine the feasibility of these patterns on WMT14 En-De and CNN/Daily Mail. 

\begin{table}[t]
\centering
\small
\begin{tabular}{lcccc}
\toprule
\multirow{2}{*}{Pattern} & \multicolumn{2}{c}{WMT14 En-De} & \multicolumn{2}{c}{CNN/Daily Mail} \\
\cmidrule(lr){2-3}
\cmidrule(lr){4-5}
& Sparsity & BLEU & Sparsity & RL  \\
\midrule
Baseline & 0.00\% & 27.59 & 0.00\% & 36.88 \\
\lzdrop{} & 46.7\% & 27.06 & 47.6\% & 36.51 \\
\addlinespace
POS Pattern & 46.7\% &  27.11 & 39.6\% & 35.57 \\
Freq Pattern & 42.1\% & 26.98 & 47.8\% & 35.67 \\
Group Pattern & 50.0\% & 26.82 & 50.0\% & 30.69 \\
\addlinespace
Inv Freq Pattern & 44.7\% & 26.42 & 39.0\% & - \\
\bottomrule
\end{tabular}
\caption{\label{tb_pattern} Sparsity and generation quality for different models on the WMT14 En-De (measured by tokenized case-sensitive BLEU) and the CNN/Daily Mail (measured by ROUGE-L or RL) test set. The sparsity rate is evaluated on test set.} 
\end{table}

Table \ref{tb_pattern} shows the results. On WMT14 En-De, Transformer using these rule-based patterns achieves comparable translation quality to \lzdrop{} (-0.24 to +0.05 BLEU) with similar sparsity rate. One interesting observation is that Transformer also works with language- and context-agnostic sparsity patterns (Freq Pattern).
The performance drop by Inv Freq Pattern (-0.64 BLEU) is in line with the information-theoretic expectation that information from frequent words is easier to compress than that of rare words.

However, note that we developed our heuristics to mimic the behaviour of \lzdrop{} for WMT14 En-De task. \lzdrop{} has the advantage that it is data-driven and task-agnostic so that we can easily apply \lzdrop{} to summarization. By contrast, these rule-based patterns discovered on translation tasks are not optimal for other tasks, which results in deteriorated performance on CNN/Daily Mail (-5.82 to -0.84 RL). In particular, Transformer suffers from the largest performance drop with the Group pattern (-5.82 RL). 
These results suggest that using rule-based sparse patterns to manually define the sparsity of encoder outputs is possible though the patterns lack generalization ability to different tasks.

\section{Conclusion}

By introducing a $L_0$-regularized neural sparsity layer (\lzdrop{}) in Transformer, we confirm that the encoder outputs are compressible to varying degrees. Pruning encoder outputs often results in a drop in performance, but we can get comparable results with 40--70\% source encodings dropped. One benefit of pruning source encodings is to shorten encoding sequences for the decoder, which accelerates the decoding speed by up to 1.65$\times$ on document summarization tasks. Our analysis on WMT14 En-De shows that \lzdrop{} learns to drop the encodings of (relatively frequent) function words and retain encodings of (relatively rare) content words, but relies on self-attention to reroute information from these to-be-pruned positions.
Based on our analysis, we define rule-based sparsity patterns, which also allow for compression without degrading translation quality much, and show that frequent tokens are more amenable to sparsification than rare tokens. However, we find that our rule-based patterns do not generalize across tasks, while \lzdrop{} is data-driven and applicable across tasks. We hope that, besides practical implication, our work contributes to better understanding encoder-decoder models.

%In the future, we aim to develop more effective sparse patterns to make Transformer more efficient. We are also interested in adapting \lzdrop{} to other sequence-to-sequence architectures and tasks.

\section*{Acknowledgments}

This project has received funding from the European Union’s Horizon 2020 Research and Innovation Programme under Grant Agreements 825460 (ELITR) and 825299 (GoURMET). Rico Sennrich acknowledges support of the Swiss National Science Foundation (MUTAMUR; no.\ 176727). Ivan Titov acknowledges support of the European Research Council (ERC StG BroadSem 678254).

\bibliography{acl2020}
\bibliographystyle{acl_natbib}

% \begin{comment}
\appendix

\section{Experimental Settings}

\paragraph{Machine Translation} WMT14 En-De and WMT18 Zh-En contain around 4.5M and 25M training sentence pairs, respectively. 

\paragraph{Document Summarization} CNN/Daily Mail pairs news articles (791 words on average) with multi-sentence summaries (63 words on average), and involves 287,227 training pairs, 13,368 validation pairs and 11,490 test pairs. WikiSum contains 1.58M training pairs, 38,144 validation pairs and 39,357 test pairs. The used parameters for ROUGE-1.5.5.pl are \textit{-m -a -n 2}.

\paragraph{Model Settings} We augment the MLE loss with label smoothing of 0.1. We use Adam optimizer ($\beta_1=0.9, \beta_2=0.98$)~\cite{kingma2014adam} for parameter tuning, and schedule the learning rate based on the inverse square root of running steps with a warm-up step of 4K. We apply dropout to attention weights and residual layers to avoid overfitting, with a rate of 0.1/0.1 except for CNN/Daily Mail where 0.3/0.5 is used. We train different models with varied training steps: 300K for WMT14 En-De, 500K for WMT18 Zh-En, 100K for WikiSum and 80K for CNN/Daily Mail, where sequence pairs of roughly 25K target subwords are organized into one minibatch. We average the last 5 checkpoints for evaluation where beam search is adopted for decoding with beam size of 4 and length penalty of 0.6. 

% \end{comment}

\end{document}